\documentclass[hyphens]{article}
\usepackage[final]{neurips_2025} 

\usepackage[utf8]{inputenc} 
\usepackage[T1]{fontenc} 
\usepackage{hyperref} 
\usepackage{subcaption}
\hypersetup{
 colorlinks=true,
 linkcolor=red,
 citecolor=cyan,
 filecolor=magenta, 
 urlcolor=magenta,
 }
\usepackage{url} 
\usepackage{booktabs} 
\usepackage{amsfonts} 
\usepackage{nicefrac} 
\usepackage{microtype} 
\usepackage[dvipsnames]{xcolor}
\usepackage{amsmath}
\usepackage{xspace} 
\usepackage{enumitem} 
\usepackage{textcomp}
\usepackage{stfloats}
\usepackage{verbatim}
\usepackage{wrapfig}
\usepackage{graphicx}
\usepackage{subcaption}
\usepackage{amssymb}
\usepackage{cite}
\usepackage{tikz}
\usepackage{algorithm}
\usepackage{algpseudocode}
\def\ie{\emph{i.e., }}

\def\eg{\emph{e.g., }}
\usepackage{multicol,multirow} 
\usepackage{threeparttable} 
\usepackage{pgfplots}

\usepackage{fancyhdr}
\renewcommand{\headwidth}{\textwidth}

\renewcommand{\headrulewidth}{0.5pt}
\renewcommand{\headrule}{\vspace{2pt}\hbox to\headwidth{\color{black}\leaders\hrule height \headrulewidth\hfill}}
\pgfplotsset{compat=1.18}

\def\modelname{X-World}
\title{X-World: Controllable Ego-Centric Multi-Camera World Models for Scalable End-to-End Driving}
\author{\colorbox{white}{\includegraphics[height=0.8em]{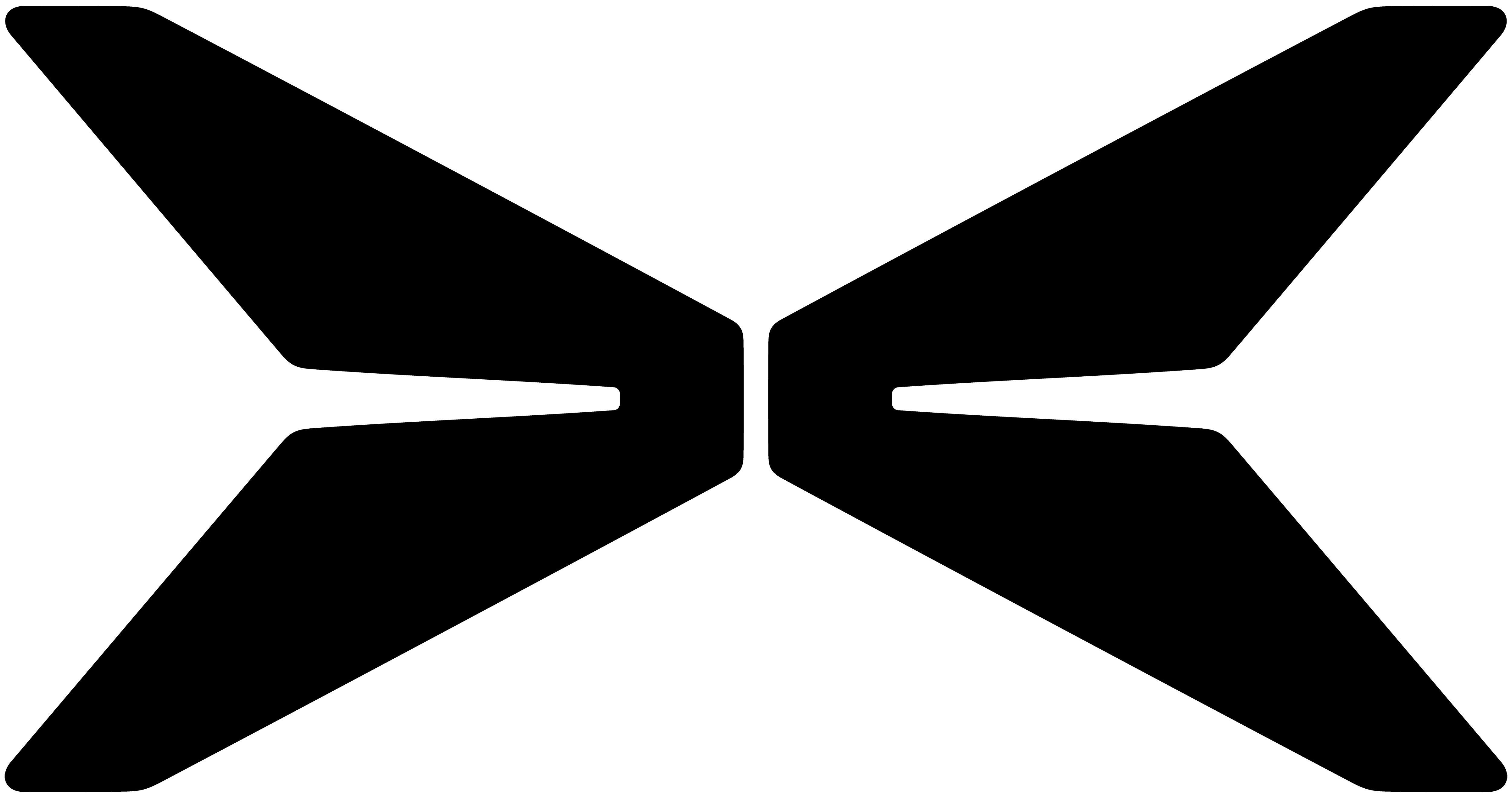}} GWM Team, XPeng Inc. \\
\textbf{\href{https://x-world-1.github.io}{\textcolor{NavyBlue}{https://x-world-1.github.io}}}}

\begin{document}
\maketitle

\begin{abstract}
Scalable and reliable evaluation is increasingly critical in the end-to-end era of autonomous driving, where vision--language--action (VLA) policies directly map raw sensor streams to driving actions. Yet, current evaluation pipelines still rely heavily on real-world road testing, which is costly, biased toward limited scenario coverage, and difficult to reproduce. These challenges motivate a \emph{real-world simulator} that can generate realistic future observations under proposed actions, while remaining controllable and stable over long horizons.
We present \textbf{\modelname}, an action-conditioned \emph{multi-camera} generative world model that simulates future observations directly in video space. Given synchronized multi-view camera history and a future action sequence, \modelname~generates future multi-camera video streams that follow the commanded actions. To ensure reproducible and editable scene rollouts, \modelname~further supports optional controls over \emph{dynamic traffic agents} and \emph{static road elements}, and retains a text-prompt interface for appearance-level control (\eg, weather and time of day). Beyond world simulation, \modelname~also enables \emph{video style transfer} by conditioning on appearance prompts while preserving the underlying action and scene dynamics. At the core of \modelname~is a multi-view latent video generator designed to explicitly encourage \emph{cross-view geometric consistency} and \emph{temporal coherence} under diverse control signals.
Experiments show that \modelname~achieves high-quality multi-view video generation with (i) strong view consistency across cameras, (ii) stable temporal dynamics over long rollouts, and (iii) high controllability with strict action following and faithful adherence to optional scene controls. 
These properties make \modelname~a practical foundation for scalable and reproducible evaluation.
Its streaming, interactive rollout interface further makes \modelname~well-suited for online reinforcement learning of end-to-end autonomous driving systems through closed-loop simulation.
\end{abstract}

\section{Introduction}
End-to-end autonomous driving is entering a new phase where learning-based systems directly map rich sensor streams to driving actions. Recent progress in end-to-end driving stacks—especially those powered by vision–language–action (VLA) models—suggests a promising path toward unified perception, prediction, and planning within a single scalable framework~\cite{intelligence2025pi05visionlanguageactionmodelopenworld,black2026pi0visionlanguageactionflowmodel,hwang2024emma}. These models benefit from large-scale data and strong representation learning, and they increasingly demonstrate competence in complex urban interactions. However, as end-to-end systems become more capable, a central bottleneck is no longer only model capacity, but the lack of a scalable and reliable evaluation and training infrastructure that matches the pace of progress.

A scalable real-world simulator is particularly critical in the end-to-end era. Unlike modular pipelines, where intermediate outputs can be unit-tested (e.g., detection AP, tracking MOTA, forecasting ADE~\cite{fong2021panoptic,nuscenes2019,WaymoWorldModel2026}), end-to-end policies are typically evaluated by closed-loop driving outcomes that are sensitive to subtle distribution shifts and compounding errors. Unfortunately, current evaluations for end-to-end systems still primarily rely on real-world testing, which is costly to run and difficult to scale. More importantly, real-world testing is often biased and under-covered—limited by geography, weather, traffic density, and the rarity of safety-critical events—and is therefore hard to reproduce and hard to compare fairly across methods. This evaluation gap slows iteration, obscures failure modes, and makes it challenging to establish trustworthy progress for end-to-end autonomy.

Beyond evaluation, an interactive real-world simulator is also a key enabler for effective online reinforcement learning (RL) for VLA systems. Online RL requires large volumes of closed-loop interactions and diverse counterfactual experiences—precisely the regimes where real-world driving is unsafe, expensive, and ethically constrained. A simulator that can faithfully model future sensor observations under proposed actions can unlock scalable training signals: policies can explore alternative maneuvers, learn to recover from off-nominal or risky states, and improve robustness under rare events, all while remaining in a controlled and repeatable environment. In this sense, simulation is not merely a testbed—it becomes the engine for continual improvement.

However, building a simulator suitable for end-to-end autonomy is challenging. It must generate photorealistic sensor observations while maintaining strict action-following (causal consistency with ego motion), cross-view consistency (agreement among multiple cameras), and long-horizon stability (avoiding drift over extended rollouts). Moreover, it should support controllability over dynamic agents and static road elements when optional conditions are available, enabling targeted scenario synthesis and stress-testing. Existing approaches often fall short along one or more of these axes: they may produce visually plausible videos but ignore fine-grained action causality, generate per-view videos that disagree geometrically, or degrade quickly when rolled out long-term. Many high-quality video diffusion models are also designed for offline, bidirectional generation over a full clip, which limits their applicability for real-time interaction.

In this work, we present \textbf{\modelname}, a controllable multi-view generative world model for autonomous driving built upon video diffusion. Given history multi-camera video streams and a driving action (or action sequence) to be applied, our model generates the corresponding future multi-camera video streams. When available, the model additionally accepts optional conditions to control dynamic traffic agents and static road elements, enabling targeted counterfactual rollouts and scenario editing. 
Besides the action and scene-element control modules, we retain a text-conditioning branch to control global appearance (\eg, weather and time of day), enabling controllable appearance-driven style editing of the generated videos.
Crucially, unlike traditional bidirectional video diffusion models, \modelname~operates in a streaming, autoregressive fashion, generating futures incrementally for real-time interaction. This design makes the model naturally compatible with closed-loop usage—both for scalable evaluation of end-to-end policies and for online RL training where the simulator must respond promptly to newly sampled actions.

Our approach achieves high-quality multi-view video generation with (i) high view-consistency, ensuring coherent geometry and object identity across cameras; (ii) strict action following, producing futures aligned with commanded ego behavior; and (iii) long video rollout ability, enabling stable predictions over extended horizons. Together, these properties move generative world models closer to a practical “real-world simulator” abstraction—one that can support reproducible benchmarking, scalable regression testing, and interactive learning for end-to-end/VLA driving systems.
\section{Data}
\subsection{Data Format}
The X-World model is trained on a meticulously curated dataset comprising a large scale of high-fidelity real-world driving sequences. These sequences are characterized by their diversity, covering a wide array of external environments, heterogeneous ego-vehicle behaviors, and complex multi-agent interactions. Each data sample constitutes a 10-second temporal segment and integrates the following multimodal data streams:

\begin{itemize}
    \item \textbf{Multiview Video Streams}: Synchronized video feeds from seven surrounding cameras.
    \item \textbf{Dynamic Object Trajectories}: Sequences of dynamic agents (e.g., vehicles, pedestrians) identified using a high-precision dynamic perception model.
    \item \textbf{Static Scene Elements}: Annotations of static infrastructure (e.g., lanes, traffic signs) obtained from a high-precision static perception model.
    \item \textbf{Textual Scene Descriptions}: Natural language descriptions of the driving scenario generated by the Vision Language Models (VLM).
\end{itemize}

\begin{figure}[htbp]
    \centering
    \begin{subfigure}[b]{0.48\linewidth}
        \centering
        \includegraphics[width=\linewidth]{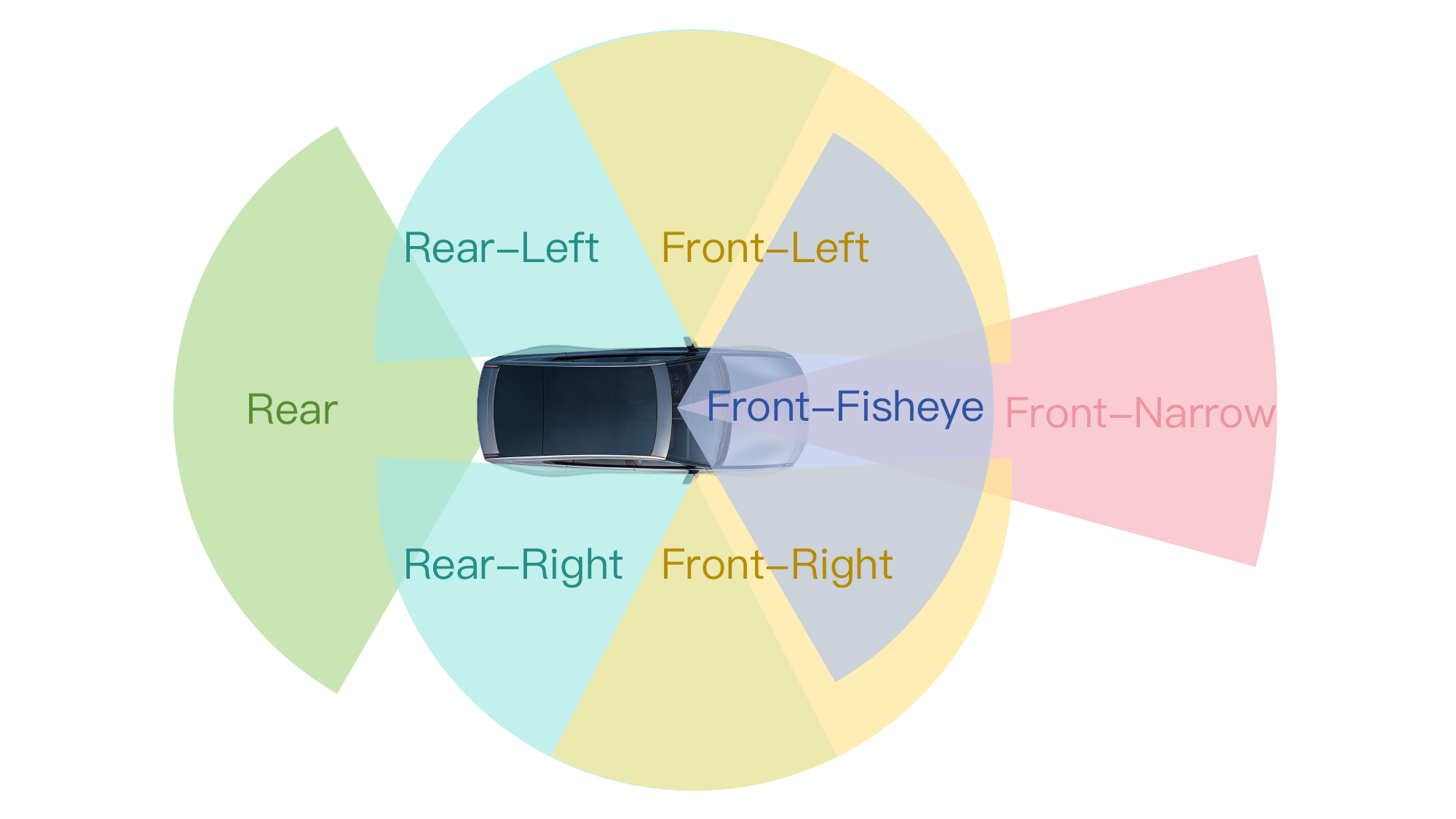}
        \label{fig:camera_layout}
    \end{subfigure}
    \hfill  
    \begin{subfigure}[b]{0.40\linewidth}
        \centering
    \includegraphics[width=\linewidth]{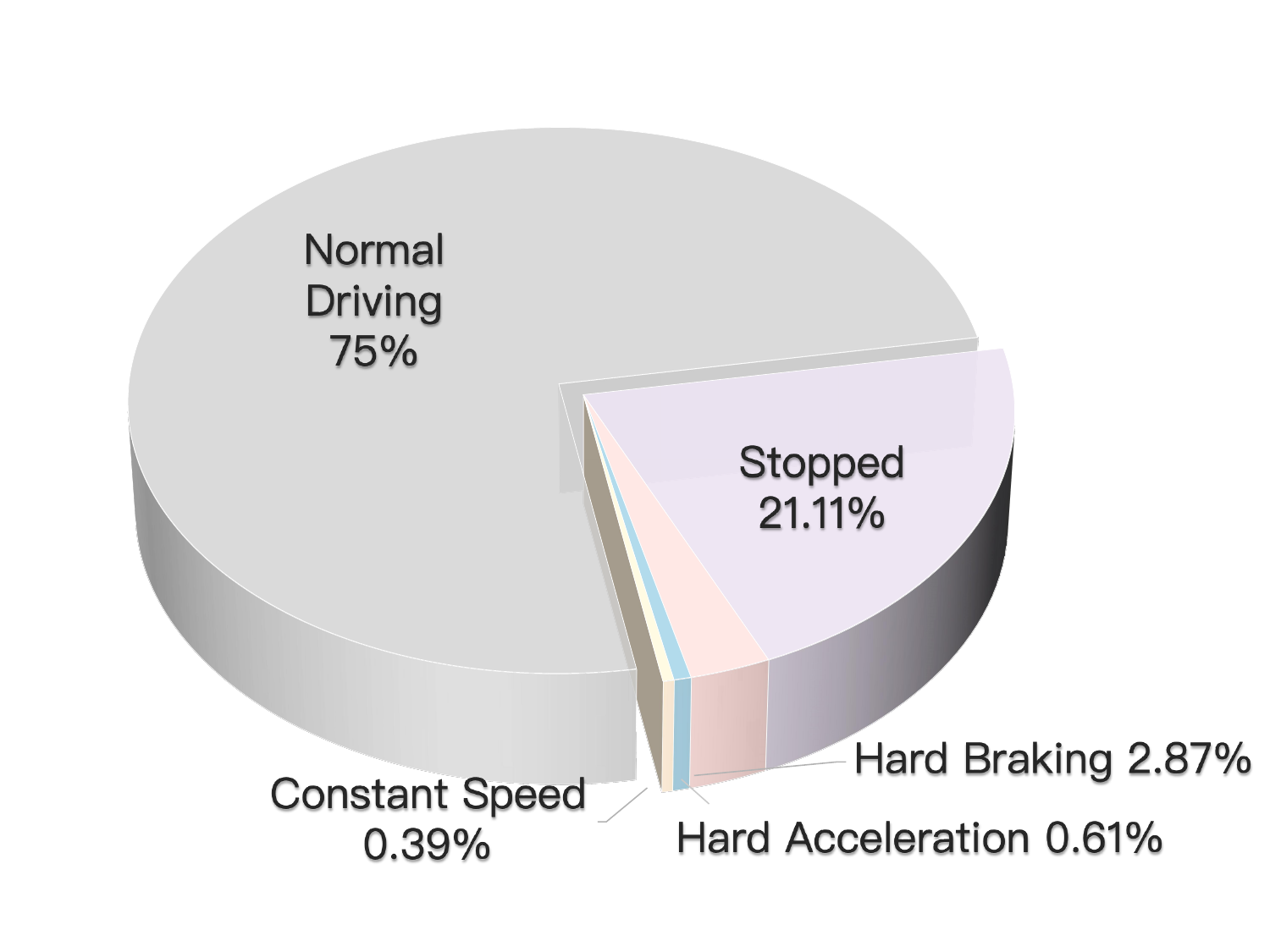}
        \label{fig:longti_action}
    \end{subfigure}
    \caption{(a) Spatial layout of the seven camera sensors. (b) Distribution of ego longitudinal actions.}
    \label{fig:combined}
\end{figure}

The video data is recorded at a rate of 12 frames per second (FPS). Each frame provides a comprehensive 360-degree surround-view through seven distinct, calibrated camera perspectives: \textit{front\_narrow}, \textit{front\_fisheye}, \textit{front\_left}, \textit{front\_right}, \textit{rear\_left}, \textit{rear\_right}, and \textit{rear}. The precise spatial configuration and field-of-view overlap of these cameras are designed to ensure full coverage around the vehicle, as illustrated in Figure~\ref{fig:combined}(a).

\subsection{Video Captioning}

To enable fine-grained control and semantic understanding in X-World, we construct a large-scale video captioning pipeline tailored for autonomous driving scenes. Unlike generic video captions, our annotations focus on driving-specific attributes that are essential for controllable scene generation and downstream evaluation. 

\noindent\textbf{Captioning Schema.} Following our quantitative evaluation protocol, each video clip is annotated across four major dimensions:

\begin{itemize}
    \item \textbf{Macro Environment:} weather (sunny, cloudy, rainy, etc.), time of day (dawn, daytime, dusk, night), lighting condition, and driving environment (region type + road type).
    \item \textbf{Road Conditions:} surface condition (flat/bumpy), slope (uphill/downhill), and road state (dry/wet/puddles).
    \item \textbf{Traffic Infrastructure:} presence of lane markings, guardrails, traffic signs, traffic lights, buildings, vegetation, and special elements (bridges, construction zones, toll stations).
    \item \textbf{Traffic Density:} five-level scale from ``empty'' to ``congested''.
\end{itemize}

\noindent\textbf{Automated Pipeline.} Given the dataset, we adopt an automated approach using VLM. For each 10-second clip, we sampled a sequence of synchronized images from all 7 cameras in each 10-second fragment. The multi-view image sequence is fed into the model with a structured prompt that encodes the captioning schema.

\noindent\textbf{Example.} Below is a typical caption generated by our pipeline:

\begin{quote}
``The video captures a sunny daytime scene on a flat urban highway with sufficient sunlight. The road is lined with tall buildings and green vegetation, marked by clear white lane lines and roadside guardrails. A pedestrian overpass spans across the road in the distance. Traffic lights and signs are visible. Traffic density is moderate.''
\end{quote}

This structured, rule-guided approach ensures that every video clip in our dataset is paired with accurate, consistent, and semantically rich textual descriptions, laying the foundation for X-World's controllable generation capabilities.

\subsection{Auto Tagging}
To understand the natural distribution of data, enable more refined control over data distribution, and facilitate rapid data selection for small-scale feature validation, we developed a comprehensive, well-organized, and fine-grained three-level label taxonomy. Based on our task requirements, we defined four major categories of labels:

\begin{itemize}
    \item \textbf{Environmental Labels}: These describe overall scene-level characteristics and include 11 subcategories: Weather, Lighting, Road Surface Status, Road Surface Type, Road Curvature, Road Gradient, Road Structure, Road Type, Traffic Condition, Lane Clarity, and Lane Quantity Status. Each subcategory contains several fine-grained labels, resulting in a total of 50 third-level labels within the environmental category.
    \item \textbf{Static Labels}: Comprising 24 third-level labels, grouped into Road Markings, Lane Lines, Road Boundaries, Traffic Signs, Signal Applies To This Lane, Traffic Lights, and Static Obstacles.
    \item \textbf{Dynamic Labels}: Focus on five types describing traffic participants.
    \item \textbf{Ego-Vehicle Behavior Labels}: Include 21 third-level labels, mainly divided into Longitudinal, Lateral, Object Interaction, Scene Interaction, and Unreasonable Behavior.
\end{itemize}

The construction of this label taxonomy primarily relies on four information sources: (i) A high-precision dynamic perception network, which mainly serves dynamic labels; (ii) A high-precision static perception network, which mainly serves static labels; (iii) A robust online pose estimation system combined with vehicle pose information derived from high-precision sensors, which mainly serves ego-vehicle behavior labels; and (iv) the general-purpose VLM, which mainly serves environmental labels.

\subsection{Data Distribution}
We invested substantial computational resources to annotate the entire training dataset, enabling a comprehensive analysis of the natural data distribution and informed adjustments to the training set based on both statistical insights and model performance. Leveraging these labels, we conducted extensive iterative experiments, extracting valuable guidance for model training. As an example, Figure \ref{fig:combined}(b) illustrates the distribution of ego-vehicle longitudinal behaviors: the vast majority are normal driving (74.8\%), followed by stationary (21.0\%), while the remaining categories constitute a long tail. This analysis directly informs data collection—for instance, if the model exhibits poor performance on hard acceleration, we prioritize acquiring more such samples to enhance overall efficacy.


\section{Method}



   

\begin{figure*}[t] 
  \centering
   \includegraphics[width=\linewidth]{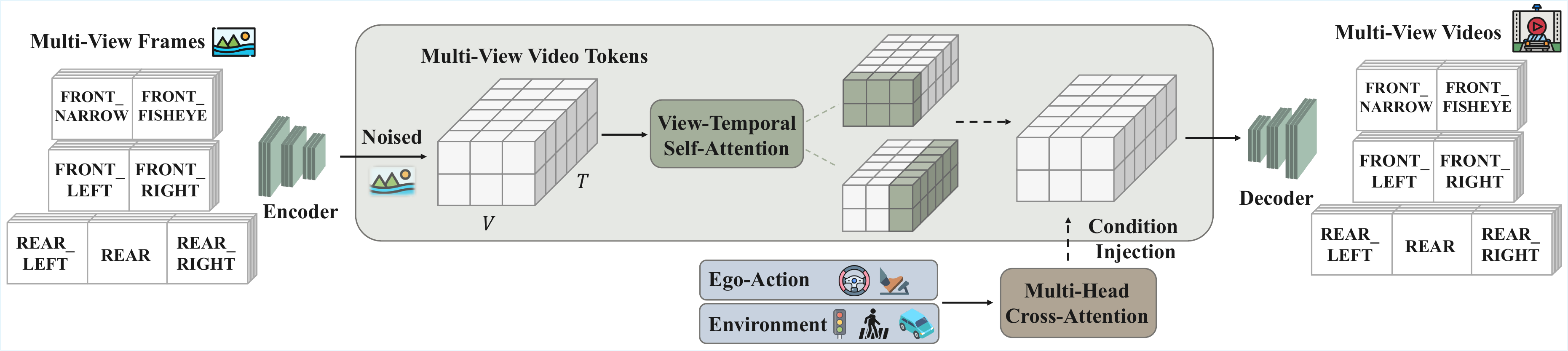} 
   \caption{The overall scheme of the proposed X-World.
   }
   \label{fig:overall}
   
\end{figure*}
\subsection{Overview}\label{sec:method_overview}
Modern embodied agents---including autonomous driving systems---primarily perceive and reason about the world through \emph{cameras}. As a result, the effective ``world state'' available to these agents is not a compact vector of latent variables, but a high-dimensional stream of images, \ie, \emph{video}. This motivates a world model that operates directly in the observation space most relevant to downstream policies: \emph{action-conditioned video}.

We propose \textbf{\modelname} as in Fig.~\ref{fig:overall}, a generative world model formulated as an \emph{action-conditioned multi-camera video generation model}. Given a short \emph{history} of synchronized multi-view camera streams, the model predicts the \emph{future} camera observations that would result from executing a specified \emph{future action sequence}. 
Concretely, our model takes as input:
(i) multi-camera video history $\mathbf{X}_{t-L:t}^{1:V}$ from $V$ cameras, representing the recent visual context of the scene;
(ii) future driving actions $\mathbf{A}_{t:t+H}$ to be executed by the ego vehicle;
and (iii) optional scene control conditions $\mathbf{C}$, which specify controllable aspects of the environment.
It then generates the corresponding multi-camera future video $\hat{\mathbf{X}}_{t+1:t+H}^{1:V}$ that is (i) visually realistic, (ii) consistent across views, and (iii) faithful to the commanded actions:
\begin{equation}
\hat{\mathbf{X}}_{t+1:t+H}^{1:V} \sim
p\!\left(\mathbf{X}_{t+1:t+H}^{1:V}\mid \mathbf{X}_{t-L:t}^{1:V}, \mathbf{A}_{t:t+H}, \mathbf{C}\right).
\label{eq:world_model}
\end{equation}

A key practical requirement for evaluation and training is \emph{reproducibility}: we often want the simulator to produce the same future (or a controlled family of futures) under specified conditions. To this end, our model optionally supports explicit control over both \emph{dynamic traffic agents} (e.g., surrounding vehicles, pedestrians) and \emph{static road elements} (e.g., lane topology, road layout cues). When such conditions $\mathbf{C}$ are provided, the model can generate scene-consistent and reproducible futures, enabling controlled counterfactual rollouts, fair benchmarking, and systematic stress testing.


\subsection{Model Design}\label{subet_bm}
\modelname~is built upon the state-of-the-art WAN~2.2~\cite{wan2025wanopenadvancedlargescale}, following its latent video generation paradigm that couples a video VAE with a DiT-based latent denoiser~\cite{peebles2023scalable}.
In particular, consistent with WAN~2.2~5B~\cite{wan2025wanopenadvancedlargescale}, we employ a high-compression 3D \emph{causal} Variational Autoencoder that achieves a $16\times$ spatial compression ratio and a $4\times$ temporal compression ratio, producing latents with channel dimension $48$.
Operating in this compact spatio-temporal latent space substantially reduces compute and memory overhead, which (i) enables pre-training on longer video sequences to better capture rich spatio-temporal dependencies, and (ii) facilitates faster inference for downstream deployment.\newline

To address the critical challenge of \emph{geometric consistency} in multi-camera autonomous driving scenarios, we introduce a customized DiT block tailored to the multi-condition generative framework of \modelname. The design has two key goals: (i) enforcing \emph{spatio-temporal modeling} with strong \emph{cross-view consistency}, and (ii) enabling \emph{controllable generation} under heterogeneous condition signals (\eg., actions, camera parameters, dynamic agents, static road elements, and text prompts) with minimal cross-condition interference.

\paragraph{View-temporal self-attention.}
At the core of our architecture is a \textbf{view-temporal self-attention} module that explicitly models interactions along both the \emph{temporal} and \emph{cross-view} dimensions. Concretely, self-attention is executed over latent tokens across multiple cameras and multiple timesteps alternately, allowing features to align and exchange information across views while maintaining temporal coherence. This mechanism encourages consistent geometry, object identity, and motion patterns among synchronized cameras.

\paragraph{Condition injection strategy.}
We employ modality-appropriate condition injection mechanisms to balance expressiveness and stability. Specifically, we use:
(i) \textbf{adaptive layer normalization} for injecting \emph{actions} and \emph{diffusion/flow timesteps};
(ii) \textbf{additive embeddings} for \emph{camera parameters};
and (iii) \textbf{cross-attention} for high-level and structured conditions, including \emph{dynamic agents}, \emph{static road elements}, and \emph{text prompts}. Details of the conditioning design are provided in Section~\ref{subsubsec_fm}.

\paragraph{Decoupled cross-attention for heterogeneous conditions.}
We adopt decoupled cross-attention layers to fuse heterogeneous condition sources in a modular manner. Rather than injecting all conditions through a single shared attention pathway, we allocate separate cross-attention branches for different modalities.
The original text-conditioning branch from WAN~2.2~5B is retained to support optional appearance and scene-level control, such as weather, daytime, and other global attributes. For dynamic and static controls, we introduce new cross-attention branches.
This decoupling reduces mutual interference between condition types and improves controllability, enabling the model to follow each condition signal more faithfully.

\subsection{Conditions} \label{subsubsec_fm}



X-World provides a comprehensive set of conditional control interfaces that enable fine-grained manipulation of the driving scene generation process. 
These include ego-vehicle actions, dynamic agents, static road elements such as lane lines and boundaries, as well as camera intrinsic and extrinsic parameters.


\paragraph{Ego-vehicle Action.}
Controlling ego-vehicle actions in a world model enables causally consistent future simulation conditioned on planned maneuvers, which is essential for closed-loop planning and safety validation. 

Unlike high-level command conditioning, our model enables direct and continuous control by taking as input a sequence of future kinematic states --- velocity, curvature, roll, and pitch. 
Given the disparate numerical scales of these four kinematic variables, we first normalize each through \textit{symlog} normalization~\cite{webber2013bi}. 
To capture fine-grained distinctions in scalar values, we subsequently apply Fourier feature encoding. 
The encoded representations are then projected and aligned to the latent space dimension using an MLP. 
Finally, we incorporate timestamp embeddings and inject the combined conditioning signals into the diffusion blocks via adaLN-Zero~\cite{peebles2023scalable}.

\paragraph{Dynamic Agents.}
Controlling dynamic agents in a world model enables the simulation of diverse, interactive traffic behaviors, which is essential for evaluating the robustness and safety of autonomous driving policies under realistic multi-agent scenarios.

To represent dynamic agents, we first extract their semantic categories (e.g., SUV, pedestrian, bicycle) and spatial coordinates from our detection model. 
Each categorical attribute is encoded via umT5 encoder~\cite{chung2023unimax}, while spatial coordinates are normalized and further processed with Fourier feature encoding to preserve fine-grained positional details. 
These heterogeneous features are then concatenated and projected to a unified feature dimension through an MLP. 
To effectively condition the generative process, the resulting agent embeddings are injected into the latent space via cross-attention layers, allowing the model to dynamically attend to relevant agent information at each denoising step. 
This design enables flexible control over the behavior and placement of multiple traffic participants.

\paragraph{Static Elements.}
Controlling static road elements (e.g., lane lines, boundaries) in a world model allows for the specification of diverse road topologies and traffic rules, which is essential for generating scene-compliant and geometrically plausible future simulations under varying environmental layouts. 

Similarly to the encoding and injection scheme used for dynamic agents, we first extract semantic categories and positional information of static road elements via a detection model. 
Category labels are encoded using umT5, while normalized position coordinates are embedded through Fourier feature encoding. 
These representations are then projected and aligned to the target feature dimension via an MLP and subsequently injected into the diffusion latent space through cross-attention layers. 
Unlike dynamic agents, however, static elements require stronger condition adherence during inference to ensure geometric and semantic consistency. 
To this end, we employ classifier-free guidance (CFG) at test time and incorporate a random dropout strategy during training. 
This design ensures that the model remains robust to varying levels of conditional control and can faithfully generate scene-consistent futures under explicit static constraints.

\paragraph{Camera Parameters.}
Controlling camera intrinsic and extrinsic parameters in a world model enables the generation of future image sequences conditioned on diverse sensor configurations and viewpoints, thereby accommodating various vehicle types and camera setups. T
his capability is essential for learning viewpoint-aware representations and evaluating planning models under heterogeneous sensor configurations in closed-loop simulation. 

Camera intrinsic and extrinsic parameters are first normalized individually, concatenated, and then transformed via an MLP for feature projection and dimension alignment. 
The resulting embedding is directly injected into the latent space through an additive conditioning module.

\subsection{I2V/V2V/C2V Unification}
\modelname~supports multiple generation modes by controlling the length of the history input during training. Let $L$ denote the number of clean history frames provided to the model. When $L{=}1$, the model operates in an \emph{image-to-video (I2V)} regime, where the first multi-camera frame anchors appearance and geometry and the model generates the subsequent future frames. When $L{>}1$, the model naturally becomes \emph{video-to-video (V2V)}, conditioned on a multi-frame observed history to generate future multi-view videos. When $L{=}0$, the model generates videos purely conditioned on the provided action and other control conditions, which we refer to as \emph{condition-to-video (C2V)}.

\paragraph{Discussion on C2V.}
C2V is a useful training side product but is not formally a world model, since it does not condition on the current observed state and thus does not model state transitions. Nevertheless, C2V is practically valuable: it enables controllable \emph{data synthesis} and appearance-driven \emph{style transfer} (e.g., changing weather or time of day) under fixed actions and scene controls, complementing the primary world-model functionality.

\smallskip


\begin{figure*}[t] 
  \centering
   \includegraphics[width=\linewidth]{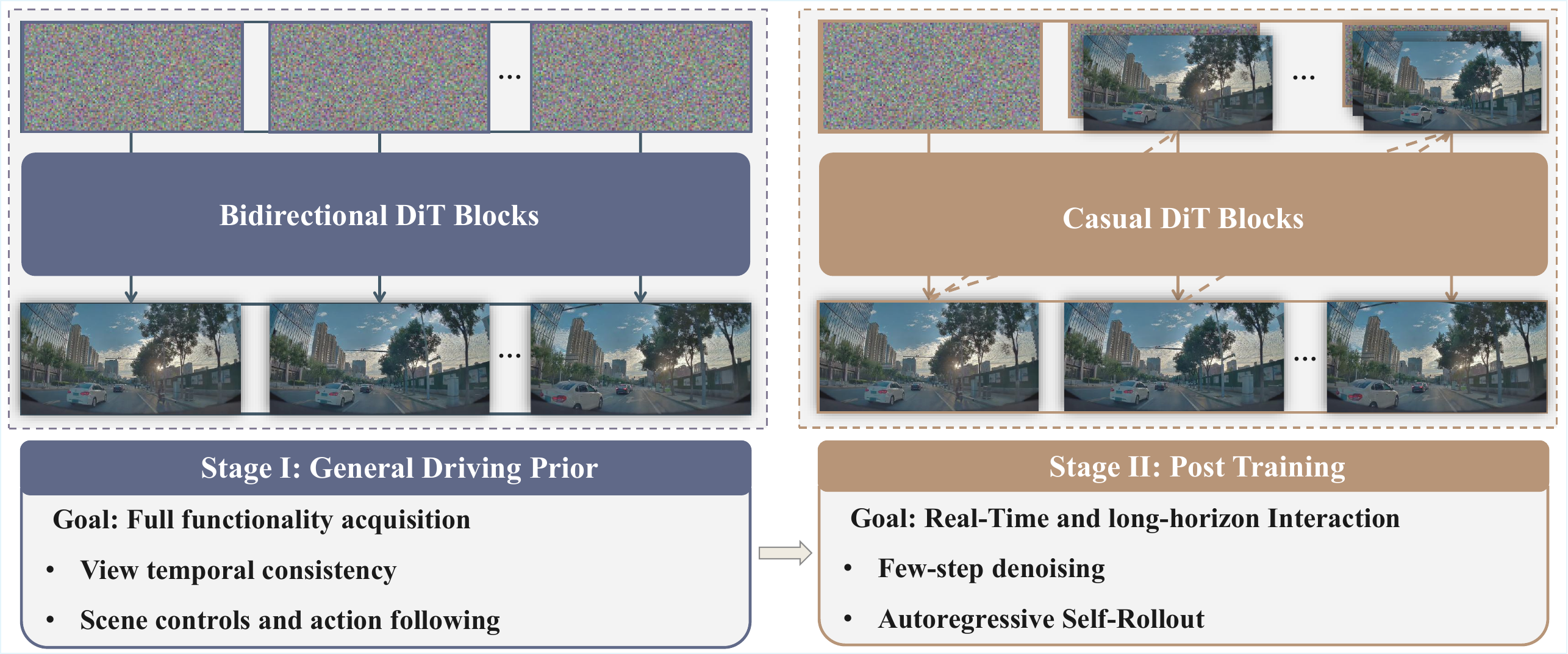} 
   \caption{The training pipeline of the proposed X-World.
   }
   \label{fig:stages}
   
\end{figure*}

\section{Training}
Our model is trained in two stages as shown in Figure~\ref{fig:stages}. \textbf{Stage-I} adapts a large pretrained video generator into a fully controllable bidirectional multi-camera world model, while \textbf{Stage-II} transforms it into a streaming autoregressive simulator for real-time interaction and long-horizon rollouts. 

\subsection{Stage-I: Bidirectional I2V Training for Accurate Controllability}
\begin{figure}[t]
  \centering
  \includegraphics[width=1.0\columnwidth]{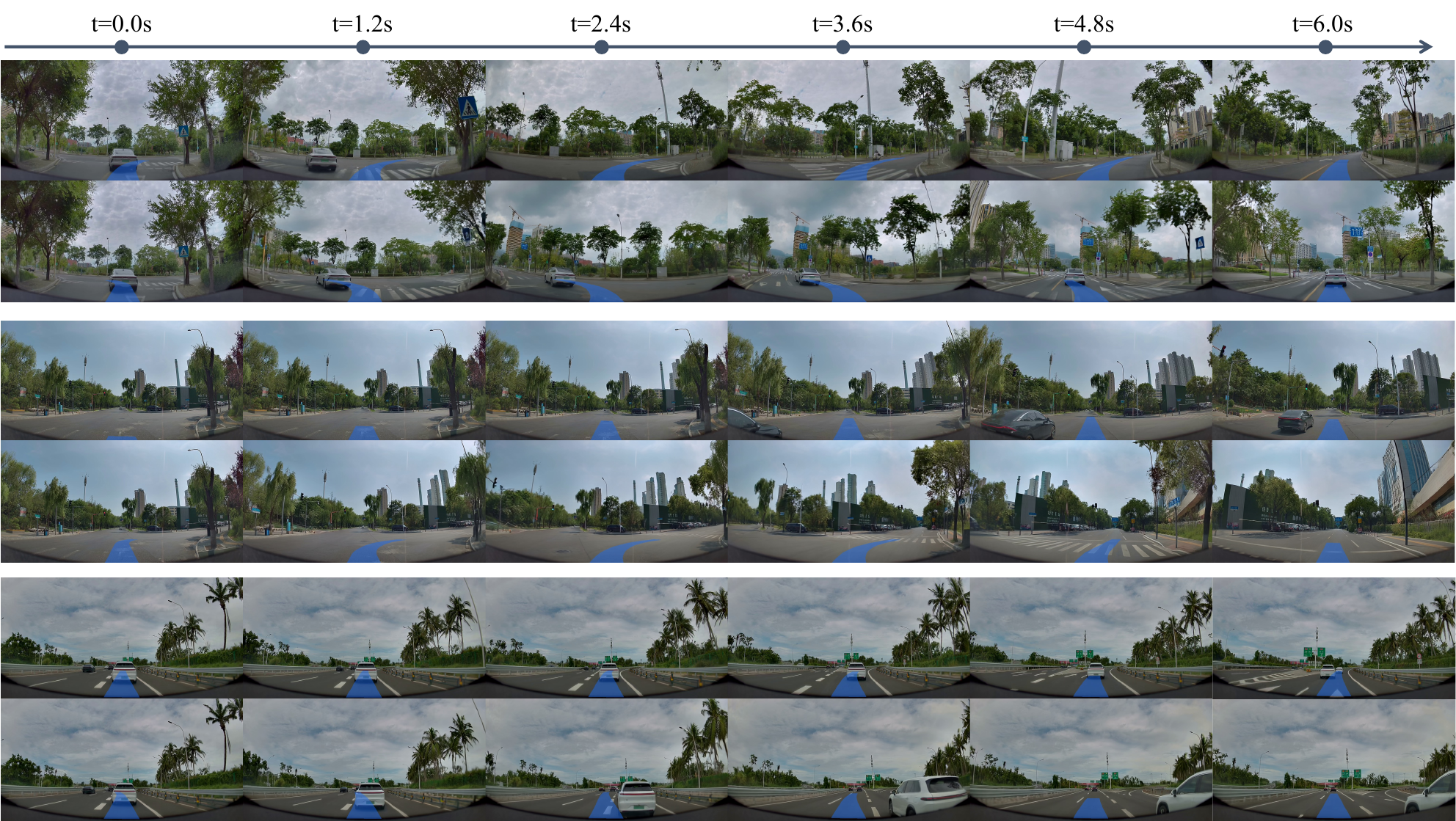}
  \caption{\textbf{Action Controllability.} Each row shows a pair of logged and generated videos. The ego trajectories for the next 3s are overlaid in {\color{NavyBlue}blue}. Dynamic traffic participants are \emph{not} constrained in this experiment and are generated freely by ~\modelname. \textbf{Top:} Turn Right $\rightarrow$ Turn Left; \textbf{Middle:} Go Straight $\rightarrow$ Turn Right; \textbf{Bottom:} Lane Keeping $\rightarrow$ Lane Change.}
  \label{fig:action_ctrl}
\end{figure}
\noindent\textbf{Initialization.}
We initialize \modelname~from WAN~2.2~5B TI2V~\cite{wan2025wanopenadvancedlargescale}. Parameters inherited from WAN are loaded directly, while newly introduced modules for our multi-camera and multi-condition setting are randomly initialized.

\noindent\textbf{Training data.}
Stage-I is trained on synchronized multi-camera short clips of \textbf{81 frames}. Each sample is paired with the corresponding driving actions and, when available, scenario-level text descriptions and structured dynamic/static control signals.

\noindent\textbf{Rectified flow objective.}
Let $\mathbf{y}$ denote the target latent video to be generated (e.g., the latent sequence of future multi-camera frames), and let $\mathbf{c}$ denote the conditioning inputs, which include the \emph{history latents} (when $L>0$) together with actions, camera parameters, optional dynamic/static controls, and text prompts. Following rectified flow~\cite{liu2022flow}, we sample $t \sim \mathcal{U}(0,1)$ and construct an interpolation between a data sample $\mathbf{y}_0 \sim p_{\text{data}}(\mathbf{y}\mid \mathbf{c})$ and Gaussian noise $\mathbf{y}_1 \sim \mathcal{N}(\mathbf{0}, \mathbf{I})$:
\begin{equation}
\mathbf{y}_t = (1-t)\mathbf{y}_0 + t \mathbf{y}_1 .
\label{eq:rf_interpolation}
\end{equation}
Rectified flow learns a time-dependent velocity field $v_\theta(\mathbf{y}_t, t, \mathbf{c})$ that matches the constant target flow $\mathbf{y}_1 - \mathbf{y}_0$ along this rectified path. The Stage-I training objective is:
\begin{equation}
\mathcal{L}_{\text{RF}}(\theta)
=
\mathbb{E}_{\mathbf{y}_0,\,\mathbf{y}_1,\,t,\,\mathbf{c}}
\left[
\left\|
v_\theta(\mathbf{y}_t, t, \mathbf{c}) - (\mathbf{y}_1 - \mathbf{y}_0)
\right\|_2^2
\right].
\label{eq:rf_loss}
\end{equation}

\noindent\textbf{Outcome and limitation.}
After Stage-I, we obtain a fully functional \textbf{bidirectional} world model that produces high-quality multi-camera futures with accurate controllability. However, similar to WAN, Stage-I relies on a \textbf{bidirectional, many-step} sampling procedure (typically $\sim$50 refinement steps for high quality) that generates an entire short clip offline, making it most suitable for short-clip synthesis rather than low-latency, long-horizon streaming rollouts.

\subsection{Stage-II: Causal Few-Step Training for Streaming Long-Horizon Simulation}

Stage-I yields a strong world model that is best suited for short-clip generation. However, it is not directly suitable for real-time interactive long-horizon rollouts because it relies on a many-step, bidirectional iterative procedure.
To address this limitation, in Stage-II, we convert it into a causal, few-step generator.
Compared to bidirectional models that generate an entire clip offline, our causal model supports streaming inference: it produces and returns future videos chunk-by-chunk, without waiting for the full sequence to finish generation. This enables low-latency interaction and naturally fits long-horizon rollouts in closed-loop settings.

\textbf{Chunk-wise causal architecture.}
A causal generator naturally supports autoregressive inference: future chunks are generated sequentially, each conditioned only on past context (history observations, previous generated chunks, and the action/scene conditions). In this setting, a KV cache further improves efficiency by reusing attention keys/values from earlier chunks, avoiding recomputation of past-context attention at every step and substantially reducing inference compute.
Following CausVid~\cite{yin2024slow}, we modify the Stage-I bidirectional model into a chunk-wise causal one. Concretely, we divide the latent sequence along the temporal dimension into contiguous chunks. Within each chunk, tokens still interact bidirectionally to preserve local spatiotemporal coherence and generation quality. However, we enforce chunk-level causality by preventing tokens from attending to any future chunks. 
As a result, the model becomes causal in time while retaining rich intra-chunk modeling capacity. 
This design provides a favorable trade-off: it enables online generation and low-latency rollouts, while avoiding the quality degradation often observed in strictly token-level causal video generation.

\begin{figure}[t]
  \centering
  \includegraphics[width=1.0\columnwidth]{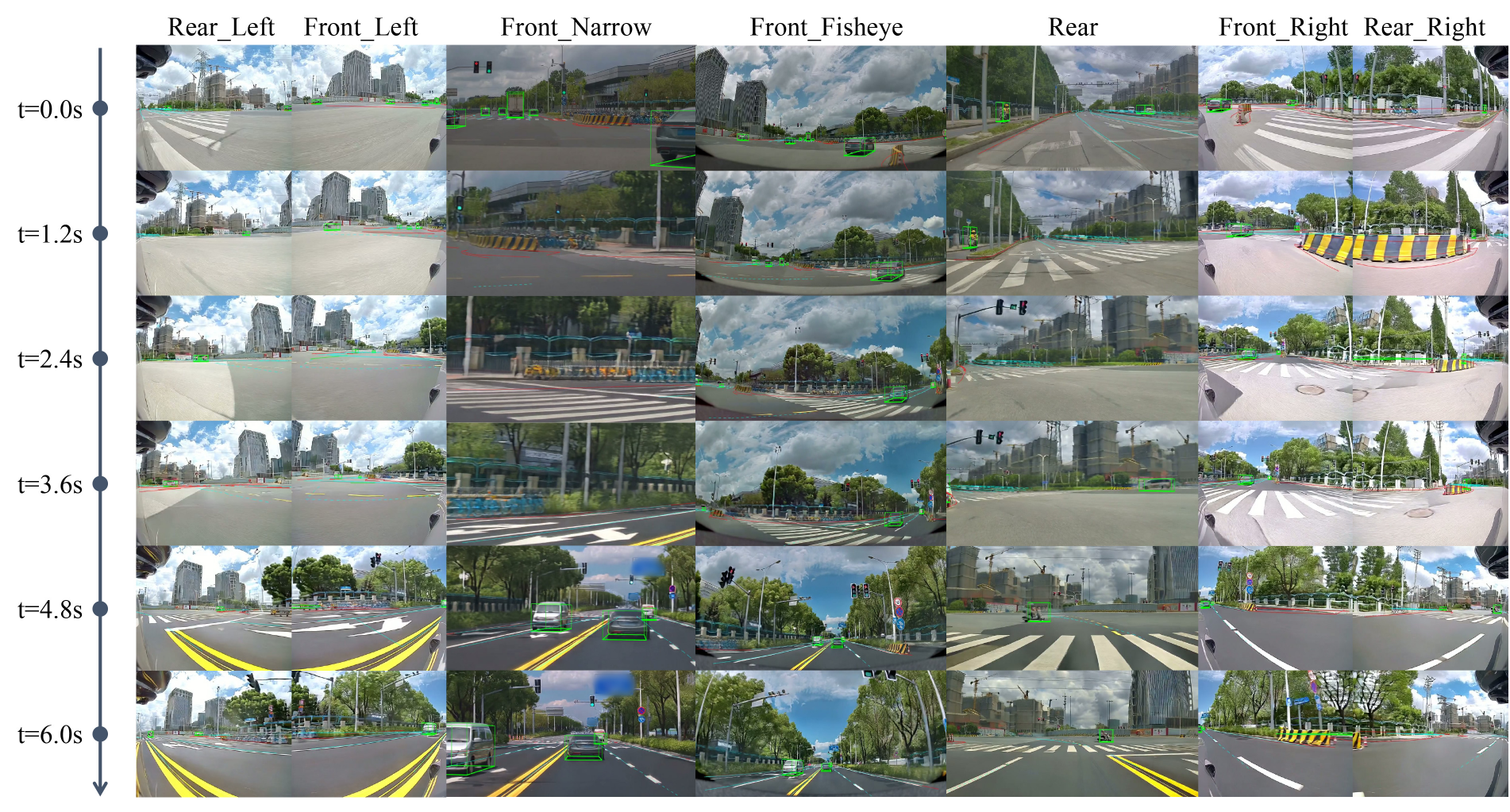}
  \caption{\textbf{Dynamic and static element controllability.} Projected bounding boxes of dynamic agents and road elements are overlaid on the videos. {\color{green}Green} denote objects. {\color{red}Red} lines denote solid road boundaries, and {\color{cyan}cyan} lines denote lane markings.}
  \label{fig:ds_ctrl}
\end{figure}

\textbf{Few-Step Self-forcing Training.}
To train the Stage-II causal generator under realistic rollout conditions, we adopt self-forcing~\cite{huang2025self}. Instead of conditioning on ground-truth history context (teacher forcing / diffusion forcing~\cite{chen2024diffusion}), the model is trained on its own autoregressive rollouts, which significantly reduces the train–test mismatch that commonly leads to compounding errors in long-horizon generation.
Concretely, generation proceeds chunk-by-chunk with KV-cache enabled for both training and inference. For each new chunk, we initialize its latent from a standard Gaussian and perform 4-step denoising, conditioned on the previously generated clean frames (together with action and optional dynamic/static conditions). This produces a self-rollout distribution induced by the Stage-II causal model.
We then optimize the model using DMD (distribution matching distillation) loss~\cite{yin2024improved,yin2024onestep}, which minimizes the reverse KL divergence between the self-rollout distribution and a target distribution represented by our Stage-I bidirectional teacher. By matching the teacher distribution under self-generated contexts, self-forcing mitigates exposure bias and reduces compounding errors in autoregressive rollouts, leading to more stable long-horizon generation. 
Moreover, since each chunk is trained to be produced with a fixed, small denoising budget, the resulting model naturally becomes a few-step generator suitable for real-time streaming simulation.

\textbf{Long video generation with Rolling KV Cache.}
Long-horizon rollouts are supported during inference using a fixed-size Rolling KV Cache. Concretely, we allocate a cache with a predetermined capacity to store the attention key/value tensors from previously generated chunks. As generation proceeds chunk-by-chunk, newly produced keys/values are appended to the cache. When the cache reaches capacity, we evict the oldest entries following a FIFO (first-in, first-out) rule, ensuring that the model always attends to a sliding window of the most recent context. This design yields bounded memory usage and stable runtime, while still providing sufficient recent temporal context for coherent long video rollouts.

Overall, Stage-II produces a causal, few-step, streaming multi-view generative world model that maintains the controllability learned in Stage-I, while enabling real-time interaction and long video generation required by scalable evaluation and online RL training for end-to-end/VLA autonomous driving systems.

\section{Results}\footnotemark
\footnotetext{Road signs, textual content, and license plates are intentionally blurred in the figures to comply with legal and privacy requirements. These modifications are applied for visualization purposes only and do not reflect the quality of either the logged data or the model-generated outputs.}
\label{sec:results}
\begin{figure}[t]
  \centering
  \includegraphics[width=1.2\columnwidth]{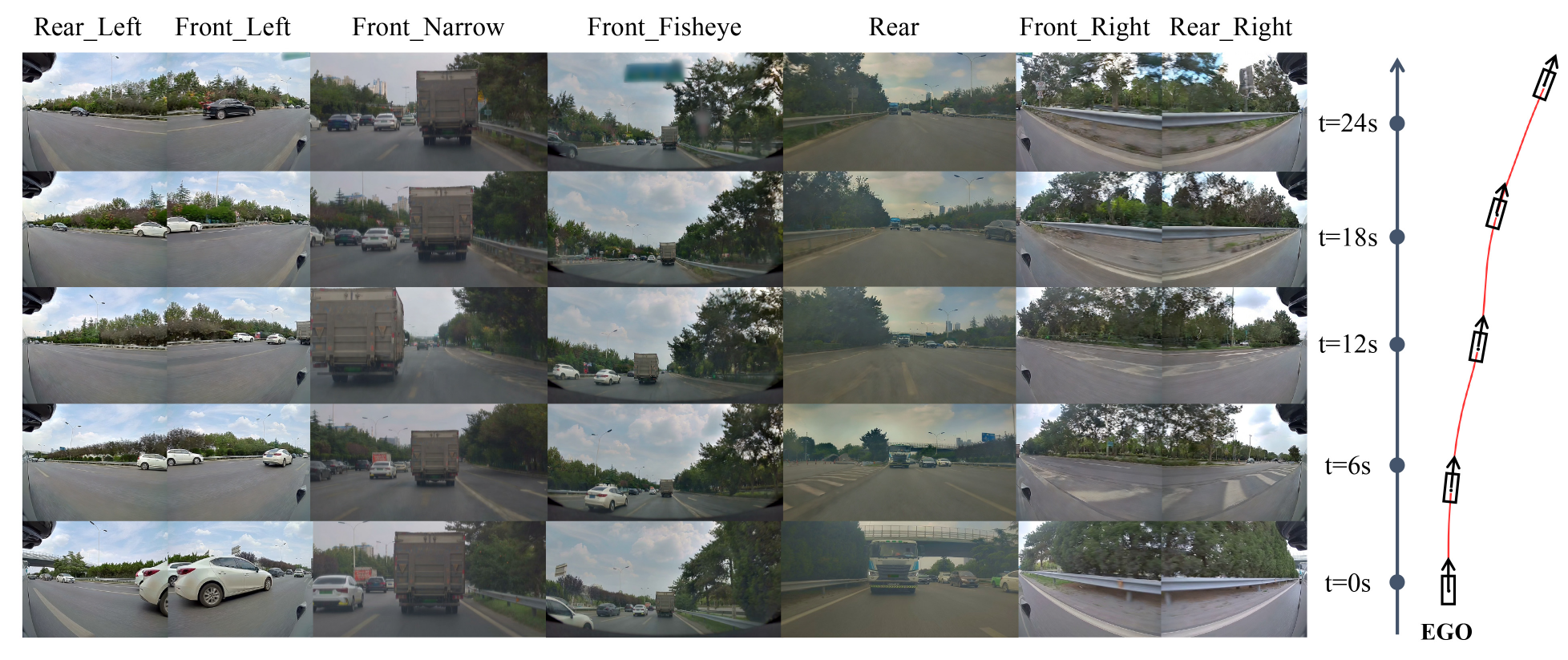}
  \caption{\textbf{Long Sequence Generation (24\,s Multi-Camera Rollout).} \modelname~produces temporally stable and view-consistent multi-camera videos over a 24\,s horizon, supporting streaming long-horizon simulation without catastrophic drift.}
  \label{fig:long_seq}
\end{figure}
This section evaluates \modelname~on its core capabilities as a controllable ego-centric multi-camera world model. We organize the results to highlight fine-grained controllability and generation quality, including: (i) ego action controllability, (ii) dynamic-agent and static-element controllability, (iii) long-horizon generation, (iv) multi-camera consistency, and (v) global appearance editing.

\subsection{Ego Action Controllability}
\label{subsec:action_ctrl}

Ego action controllability measures whether the generated future videos faithfully reflect the consequences of a given ego action sequence under a fixed initial scene context. As shown in Figure~\ref{fig:action_ctrl}, \modelname~produces photorealistic and physically plausible futures that closely follow the commanded actions. Conditioned on the same initial frame, changing the ego action sequence leads to correspondingly different, coherent rollouts, including smooth turns and lane changes. These results indicate that \modelname~internalizes a robust action-conditioned state transition in video space, enabling reactive and counterfactual rollouts for planning and policy evaluation.

\subsection{Dynamic and Static Element Controllability}
\label{subsec:ds_ctrl}

Beyond ego control, \modelname~supports fine-grained controllability over both \textbf{dynamic traffic agents} and \textbf{static road elements}, which is crucial for reproducible simulation and targeted stress testing. Figure~\ref{fig:ds_ctrl} visualizes this capability on \textbf{6\,s multi-camera} generations by overlaying projected conditions on the generated videos: {\color{green}green} boxes denote dynamic objects (\eg, vehicles), while {\color{red}red} and {\color{cyan}cyan} lines denote road boundaries and lane markings, respectively. The model preserves the specified agent locations/motions and anchors static layout cues consistently over time despite ego motion and viewpoint changes across different camera types (\eg, fisheye and narrow views). This demonstrates that dynamic and static controls can be enforced reliably and simultaneously, providing a controllable simulator interface for scenario editing and reproducible benchmarking.

\subsection{Long-Horizon Multi-Camera Generation}
\label{subsec:long_seq}

Figure~\ref{fig:long_seq} demonstrates \textbf{24\,s} long-horizon rollouts of \modelname~on multi-camera streams. The generated videos remain stable over extended horizons, maintaining coherent motion and appearance without catastrophic drift. Importantly, \modelname~can roll out beyond the short-clip setting used in Stage-I training, making it suitable for interactive simulation and closed-loop evaluation.

\subsection{Multi-Camera Consistency}
\label{subsec:mv_consistency}

Multi-camera world simulation requires both temporal coherence and cross-view alignment at each timestep. As illustrated in Figure~\ref{fig:long_seq} and Figure~\ref{fig:ds_ctrl}, \modelname~maintains strong \textbf{cross-view consistency}: dynamic objects exhibit consistent identity and motion across cameras, and static structures (\eg, lane topology and road boundaries) remain geometrically aligned across views. These qualitative results validate the effectiveness of our view--temporal modeling design for multi-camera generation.

\subsection{Global Appearance Editing via Text Prompts}
\label{subsec:style_transfer}

\begin{figure}[t]
  \centering
  \includegraphics[width=1.0\columnwidth]{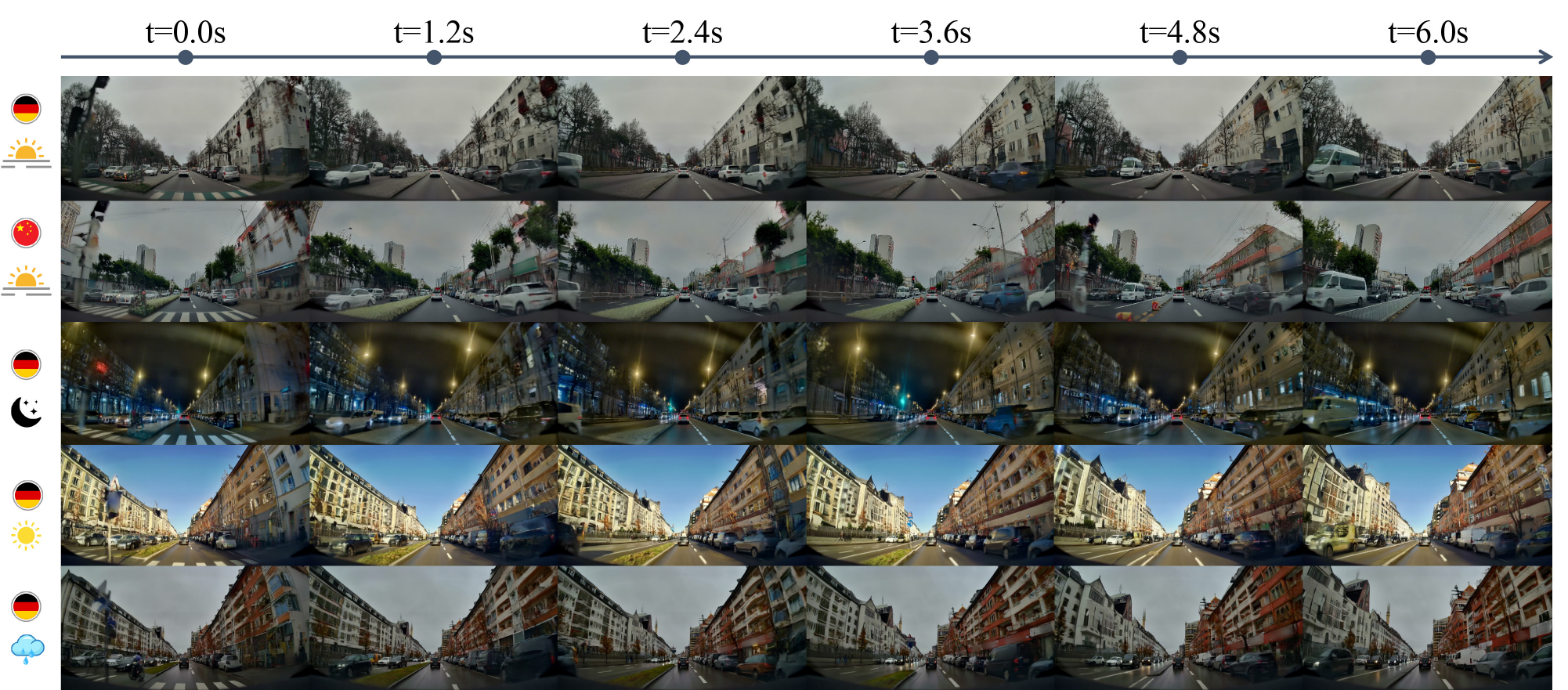}
  \caption{\textbf{C2V for Scene Appearance Editing.} Keeping the ego actions and the dynamic/static elements fixed, we vary the text prompt to edit global appearance, including locale (\eg, Germany vs.\ China), time of day (sunset vs.\ night), and weather (sunny vs.\ rainy). All images are generated by \modelname, including the first frame.}
  \label{fig:style_transfer}
\end{figure}

Finally, \modelname~supports controllable appearance editing through its text-conditioning branch, which primarily modulates global style while actions and scene elements are governed by their dedicated control branches. Figure~\ref{fig:style_transfer} shows C2V generations from the \textbf{front camera} where we keep ego actions and dynamic/static controls fixed, but vary the text prompt to edit global appearance, including locale (\eg, Germany vs.\ China), time of day (sunset vs.\ night), and weather (sunny vs.\ rainy). Notably, all frames are generated by \modelname, including the first frame, highlighting its ability to perform appearance-driven style transfer and controllable data synthesis.


\section{Applications}
In this section, we delineate the practical deployment of X-World within XPeng's autonomous driving stack. Beyond simple video generation, X-World serves as a high-fidelity, reactive, and controllable substrate for the development and validation of our next-generation Vision-Language-Action (VLA 2.0) policies.

\subsection{Closed-loop Evaluation Engine for VLA 2.0}
Although traditional 3DGS-based simulation evaluations can accurately reproduce the driving trajectories of the end-to-end driving model, they are unable to handle scenarios where the autonomous driving model makes significant lane changes, or where the running trajectory is completely different from the real vehicle's collected logs. X-World functions as a generative simulator that enables full closed-loop testing for VLA 2.0.

\textbf{Reactive Rollouts}: Unlike static log-replays, X-World responds to the ego-vehicle's real-time planned trajectories. If VLA 2.0 executes a sudden braking or steering maneuver, X-World updates the future multi-view observations accordingly, maintaining temporal and causal consistency. 

\textbf{Safety Critical Metrics}: By running VLA 2.0 within X-World, we can measure high-level performance indicators such as collision rates, progress-to-goal, and ride comfort in a virtual environment that closely mirrors the real-world visual distribution.

\begin{figure}[t]
  \centering
  \includegraphics[width=1.0\columnwidth]{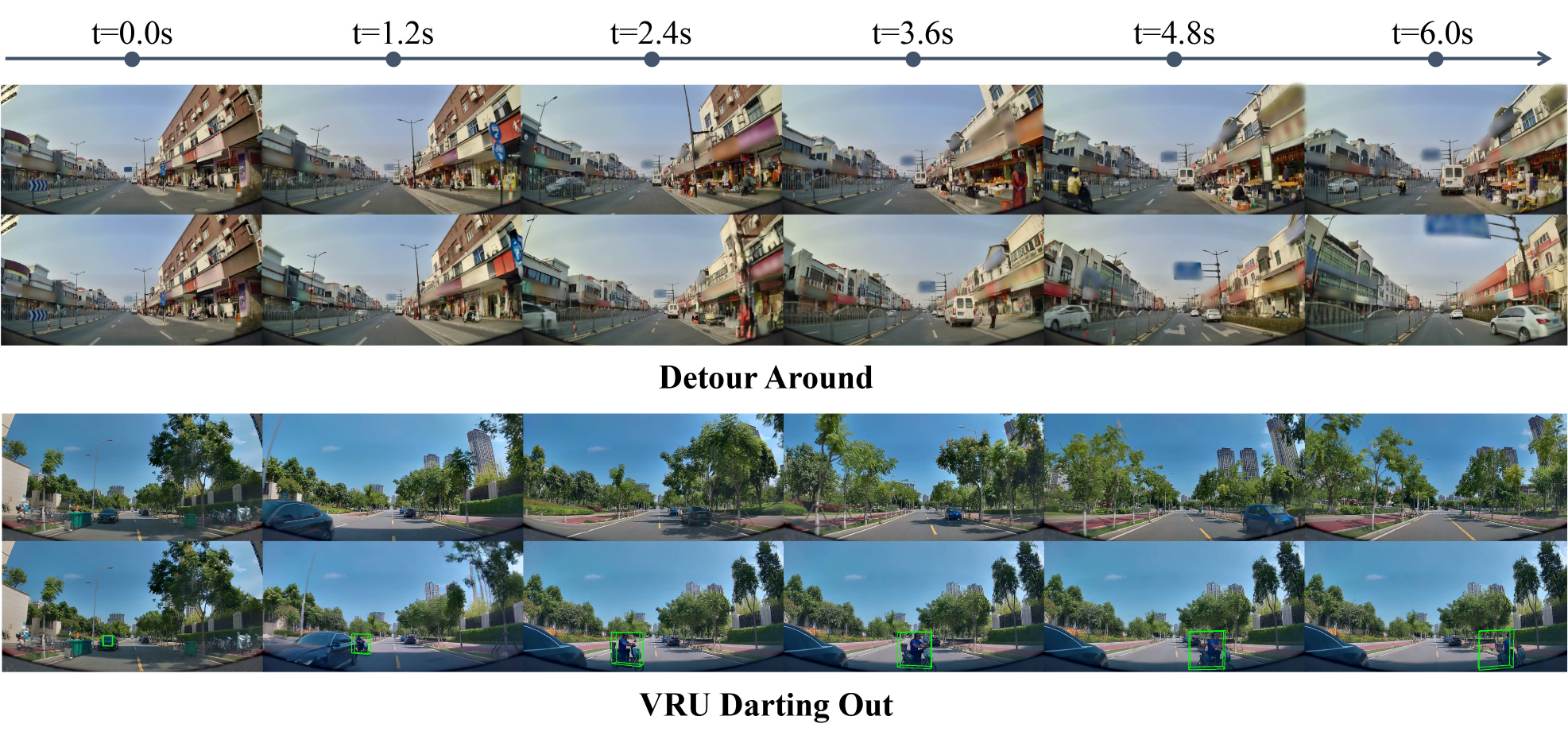}
  \caption{\textbf{Closed-loop simulation and counterfactual testing with \modelname.}
  \textbf{Scenario~1:} In the logged video, the ego vehicle waits behind a front car that is actually parked. Under the same initial scene, \modelname~enables counterfactual rollout to evaluate a policy that detours around the parked car.
  \textbf{Scenario~2:} Starting from a logged straight-driving scenario, we edit the scene by inserting a cyclist darting out from behind an occluding black car. \modelname~generates realistic futures under this edit, and the tested policy safely stops before the cyclist to avoid collision.}
  \label{fig:closed_loop}
\end{figure}

Figure~\ref{fig:closed_loop} illustrates the potential of \modelname~as a closed-loop simulator for policy evaluation under \emph{counterfactual} and \emph{edited} scenarios. We show two representative cases.

\textbf{Scenario~1 (counterfactual action rollout).}
In the logged video, the ego vehicle chooses to wait behind a front vehicle that is in fact parked. Under the same initial scene, we use \modelname~to roll out a counterfactual future conditioned on an alternative policy action: the tested policy model decides to \emph{detour} around the parked car. \modelname~generates a coherent multi-camera future consistent with this maneuver, enabling scalable evaluation of whether the policy can take the more efficient action while remaining safe.

\textbf{Scenario~2 (scene editing for safety-critical stress test).}
In the logged video, the ego vehicle goes straight and passes a nearby black car on the front-left. We then edit the scene by inserting a cyclist that darts out from behind the black car, initially occluded by it. Under this edited condition, \modelname~generates high-quality futures with consistent occlusion and motion, and the tested policy model successfully stops before the cyclist, yielding to it and safely avoiding a collision.

Together, these examples demonstrate that \modelname~can support \emph{closed-loop} evaluation by (i) rolling out alternative ego actions under the same scene and (ii) generating realistic, safety-critical counterfactuals through controllable scene editing, providing a practical testbed for end-to-end/VLA policy development.

\subsection{Simulator for Online Reinforcement Learning}
To bridge the gap between imitation learning and expert-level performance, X-World is utilized as a training environment for Online Reinforcement Learning (RL).

\textbf{Hard-case Specialization}: We leverage X-World’s controllability to stress-test VLA 2.0 in scenarios where it typically underperforms, such as "hidden-person" (ghost-outs) at intersections or indecisive lane-changing in dense traffic.

\textbf{Efficient Exploration}: By fine-tuning the policy within X-World, the VLA model can explore diverse action sequences and receive immediate visual feedback. This iterative loop allows the model to learn recovery behaviors from near-accident states—scenarios that are too dangerous to explore in the real world.

\subsection{Large-scale Data Synthesis and Augmentation}
X-World acts as a generative data factory, synthesizing rare and high-value data assets that are difficult to collect via fleet vehicles.

\textbf{Corner Case Generation}: We can procedurally generate safety-critical events, such as extreme weather conditions, rare vehicle types, or erratic pedestrian behaviors, providing a balanced training distribution that mitigates the long-tail problem.

\textbf{Overseas Expansion}: To support our global strategy, X-World enables "Zero-shot style transfer" for data. By conditioning the model on localized appearance prompts (e.g., European road markings, unique traffic signs, or left-hand traffic logic), we can transform domestic driving data into overseas training assets, significantly accelerating our international deployment without the need for extensive local data collection.


\section{Conclusion}
\label{sec:conclusion}

We presented \modelname, a controllable \emph{multi-camera} generative world model for autonomous driving that simulates future observations directly in video space. Conditioned on multi-view visual history and a future action sequence, \modelname~generates high-quality future multi-camera videos with strong \emph{cross-view geometric consistency}, \emph{temporal coherence}, and \emph{strict action following}. To support reproducible and editable scene rollouts, \modelname~additionally accepts optional controls over \emph{dynamic traffic agents} and \emph{static road elements}. In addition, we retain a text-conditioning branch that primarily modulates global appearance (\eg, weather and time of day), enabling controllable appearance-driven style editing and data synthesis while the specified actions and scene factors are enforced by their dedicated conditioning branches.

Architecturally, \modelname~builds on a latent video generation framework with a high-compression 3D causal VAE and a customized multi-view DiT backbone. The view--temporal self-attention explicitly aligns features across time and cameras, and decoupled cross-attention branches effectively fuse heterogeneous conditions with reduced interference, together enabling controllable multi-view generation under diverse inputs. To make the model practical as a ``real-world simulator'' for the end-to-end era, we further adopt a streaming, autoregressive generation interface with efficient long-horizon rollout supported by a fixed-size rolling KV cache.

Overall, \modelname~moves generative world models closer to a scalable, interactive simulator abstraction that can support reproducible benchmarking, scenario editing, and closed-loop rollouts for end-to-end/VLA autonomous driving systems, as well as controllable data synthesis and appearance-driven style transfer. We believe this direction will play an increasingly important role in scalable evaluation and training---including online reinforcement learning---by enabling low-cost, repeatable, and diverse interactions beyond what is feasible with real-world testing alone.
\newpage
\section*{Contributors}
\newcommand{\equalIT}{\textsuperscript{*}}
\newcommand{\internIT}{\textsuperscript{\ensuremath{\ddagger}}} 

We extend our sincere gratitude to the entire team for their dedication and hard work. This project is a testament to our collective effort in pushing the boundaries of world model research and engineering.

\vspace{1em}

\begin{description}
    \item[Advisors:] Yu Zhang, Xianming Liu
    \item[Project Lead:] Boyang Wang
    \item[Contributors:] Chaoda Zheng\equalIT, Sean Li\equalIT, Jinhao Deng, Zhennan Wang, Shijia Chen, Liqiang Xiao, Ziheng Chi, Hongbin Lin\internIT, Kangjie Chen\internIT
\end{description}


{\footnotesize
\noindent \equalIT Core contribution. The first two authors are listed in alphabetical order. \\
\internIT Research Intern at XPENG.}

{ 
\small
\bibliography{neurips_2025}
\bibliographystyle{plain} 
}

\end{document}